\documentclass{article} % For LaTeX2e
\usepackage[numbers]{natbib}
\usepackage{iclr2020_conference,times}

\usepackage{amsmath,amsfonts,bm}

\def\eqref#1{equation~\ref{#1}}
\def\1{\bm{1}}

\DeclareMathAlphabet{\mathsfit}{\encodingdefault}{\sfdefault}{m}{sl}
\SetMathAlphabet{\mathsfit}{bold}{\encodingdefault}{\sfdefault}{bx}{n}

\usepackage[utf8]{inputenc} % allow utf-8 input
\usepackage[T1]{fontenc}    % use 8-bit T1 fonts
\usepackage{hyperref}       % hyperlinks
\usepackage{url}            % simple URL typesetting
\usepackage{booktabs}       % professional-quality tables
\usepackage{amsfonts}       % blackboard math symbols
\usepackage{nicefrac}       % compact symbols for 1/2, etc.
\usepackage{microtype}      % microtypography
\usepackage{multirow}
\usepackage{graphicx}
\usepackage{amsmath}
\usepackage{multirow}
\usepackage{caption}
\usepackage{subcaption}
\usepackage[toc,page]{appendix}

\title{Improving Irregularly Sampled Time Series Learning with Dense Descriptors of Time}

\iclrfinalcopy

\author{
  Rafael T. Sousa\\
  Universidade Federal de Goiás\\
\texttt{rafaelts777@gmail.com}\\
  \And
  Lucas A. Pereira\\
  Universidade Federal de Goiás\\
  \texttt{apereiral@outlook.com}\\
  \And
  Anderson S. Soares\\
  Universidade Federal de Goiás\\
  \texttt{engsoares@gmail.com}\\
}

\begin{document}

\maketitle

\begin{abstract}
Supervised learning with irregularly sampled time series have been a challenge to Machine Learning methods due to the obstacle of dealing with irregular time intervals. Some papers introduced recently recurrent neural network models that deals with irregularity, but most of them rely on complex mechanisms to achieve a better performance. This work propose a novel method to represent timestamps (hours or dates) as dense vectors using sinusoidal functions, called Time Embeddings. As a data input method it and can be applied to most machine learning models. The method was evaluated with two predictive tasks from MIMIC III, a dataset of irregularly sampled time series of electronic health records. Our tests showed an improvement to LSTM-based and classical machine learning models, specially with very irregular data.

\end{abstract}

\section{Introduction}

An irregularly (or unevenly) sampled time series is a sequence of samples with irregular time intervals between observations. This class of data add a time sparsity factor when the intervals between observations are large. Most machine learning methods do not have time comprehension, this means they only consider observation order. This makes it harder to learn time dependencies found in time series problems. To solve this problem recent work propose models that are able to deal with such irregularity \citep{lipton2016modeling, bahadori2019temporal, che2018recurrent, shukla2018interpolation}, but they often rely on complex mechanisms to represent irregularity or to impute missing data.

In this paper, we introduce a novel way to represent time as a dense vector representation, which is able to improve the expressiveness of irregularly sampled data, we call it Time Embeddings (TEs). The proposed method is based on sinusoidal functions discretized to create a continuous representation of time. TEs can make a model capable of estimating time intervals between observations, and they do so without the addition of any trainable parameters.

We evaluate the method with a publicly available real-world dataset of irregularly sampled electronic health records called MIMIC-III \citep{johnson2016mimic}. The tests were made with two tasks: a classification task (in-hospital mortality prediction) and a regression (length of stay).

To evaluate the impact of time representation in the data expressiveness we used LSTM and Self-Attentive LSTM models. Both are common RNN models that have been reported to achieve great performance in several time series classification problems, and specifically with the MIMIC-III dataset \citep{lipton2016modeling, shukla2018interpolation, bahadori2019temporal, zhang2018patient2vec}. We also evaluated simpler models such as linear and logistic regression and a shallow Multi Layer Perceptron. All models were evaluated with and without TEs to asses possible improvements.

\section{Related Work}

The problem in focus of this work is how can a machine learning method learn representations from irregularly sampled data. Irregularity is found in many different areas, as electronic health records \citep{yadav2018mining}, climate science \citep{schulz1997spectrum}, ecology \citep{clark2004population}, and astronomy \citep{scargle1982studies}.

Some works deal with irregularity as a missing data problem. With time axis discretization into fixed non-overlapping intervals, those with no observations are then said to contain missing values. This approach was taken by \cite {marlin2012unsupervised}, and \cite{lipton2016modeling}. Lipton showed how binary indicators of missingness and observation time delta can improve Recurrent Neural Network based models better than imputation, even with the sparsity of binary masks.

Despite the improvement an issue about these methods is missing the potential of how the observation time can be informative \cite{little2019statistical}. More recently  \cite{shukla2018interpolation} introduced a neural network model capable to learning how to interpolate missing data and avoid time discretization, by turning a irregularly sampled time series into a regular one. \cite{bahadori2019temporal} also proposed a method to improve discretization by doing a data augmentation based on temporal-clustering.

Another approach is to make complex models capable of dealing with irregularities. The work of \cite{che2018recurrent} describes a GRU (Gated Recurrent Unit) model called GRU-D. It makes use of binary missing indicators and observation time delta as an input data and incorporates them into GRU gates to control a decay rate of missing data. \cite{bangphased} proposed a similar method using LSTM cell states to improve the decay concept.

The concept proposed in this work is similar to \cite{lipton2016modeling} and \cite{che2018recurrent}, as they propose the use of an additional input to describe observation time deltas. But instead of using time intervals we are proposing a way to describe the exact time moment using continuous cyclic functions. This way it is possible to calculate with a linear operation the time between any two irregular observations without the need of a cumulative sum over all intermediate data, also avoiding fixed-length time discretization and interpolation noise. Another difference is that Time Embeddings are dense representations that avoid unnecessary sparsity from missingness masking.

\section{Methodology}

\subsection{Positional Embeddings}

The concept of Positional Embeddings (PE) was first introduced at \cite{DBLP:journals/corr/GehringAGYD17} where the author used vectors to represent word positions in a sentence. It was initially proposed to improve Convolutional Neural Network ability to handle temporal data. As a CNN do not consider order the PE was introducing numerical representation of order into embedding latent space. 

The Transformer \citep{DBLP:journals/corr/VaswaniSPUJGKP17} network brought back the PE to improve a neural network based only on attention modules. The model had the same issue with order modeling as it contains no recurrence, they propose a set of sinusoidal functions discretized by each relative input position.

\begin{align}
\begin{aligned}
\label{TE_1}
  PE_{(pos,2i)} = sin(pos/10000^{2i/d_{model}})
\end{aligned}\\
\begin{aligned}
\label{TE_2}
  PE_{(pos,2i+1)} = cos(pos/10000^{2i/d_{model}})
\end{aligned}
\end{align}

The equations described above have a $pos$ variable to indicate position, $i$ the dimension and $d_{model}$ is the dimension of original embedding space. This way each dimension corresponds to a sinusoidal and the model is able to learn relative positions, as argued by the authors "since for any fixed offset $k$, $PE_{pos + k)}$ can be represented as a linear function of $PE_{pos)}$" \citep{DBLP:journals/corr/VaswaniSPUJGKP17}. The total dimensions of the positional embedding is defined by $d_{model}$. Each wavelength form a geometric progression from $2\pi$ to $10000 \cdot \pi$. The biggest wavelength defines the maximum number of inputs, if a position is higher than $10000$, it will start to be redundant.

\subsection{Time Embeddings}

Inspired by Transformer position representation we propose a positional embedding for irregular positions. As \cite{DBLP:journals/corr/VaswaniSPUJGKP17} discretize sinusoidal functions based on positions, it is possible to discretize it based on irregular hour times or dates. Applying these time descriptors to a irregularly sample series can make the own data be time representative.

\begin{figure}[htb]
\centering
    \includegraphics[width=0.9\linewidth]{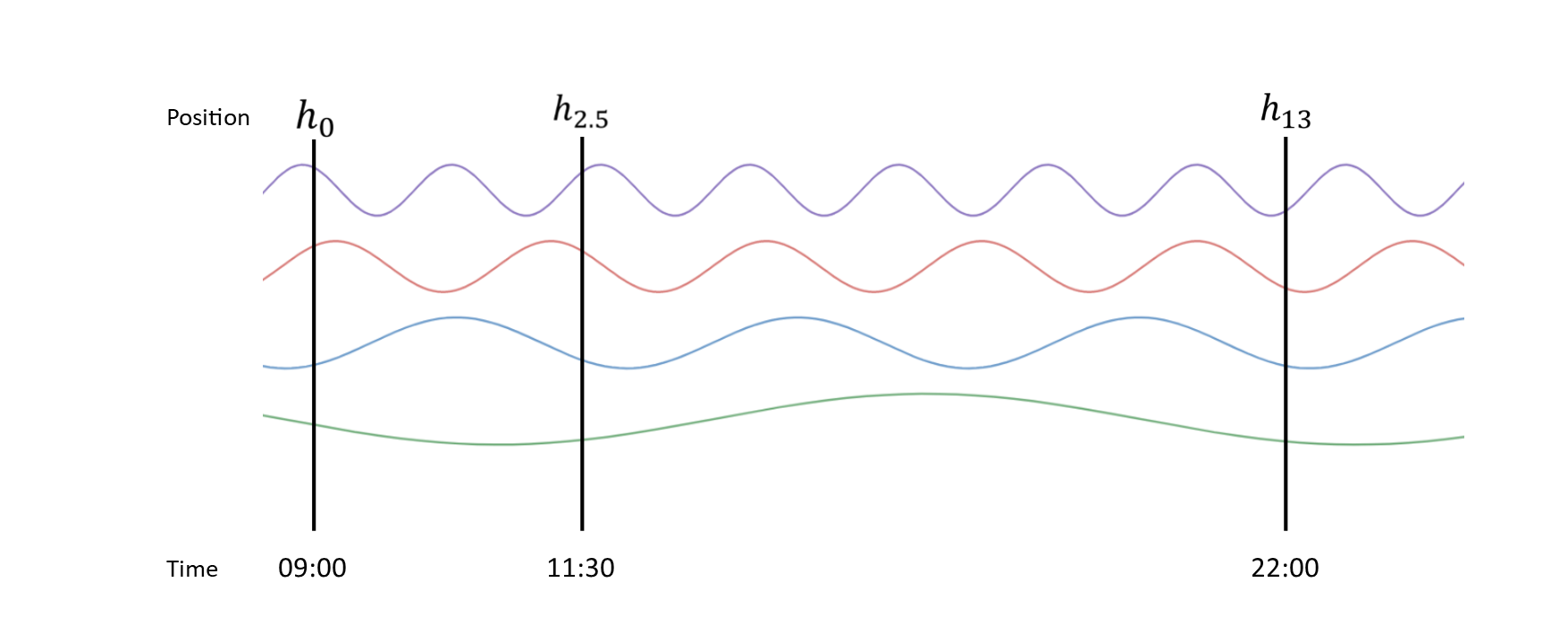}
    \centering
    \captionof{figure}{Sinusoidal functions discretization by time positions}
\label{sin_discret}
\end{figure}

\begin{align}
\begin{aligned}
\label{TE_1_2}
  TE_{(time,2i)} = sin(time/maxtime^{2i/d_{TE}})
\end{aligned}\\
\begin{aligned}
\label{TE_2_2}
  TE_{(time,2i+1)} = cos(time/maxtime^{2i/d_{TE}})
\end{aligned}
\end{align}

To do it we redefine the equations based on irregular timestamps. Instead of a position indicator there is a $time$ variable, which is continuous. The dimension of TEs ($d_{TE}$) can parameterized and a $maxtime$ defines a maximum time that can be represented. 

The relation between maximum time and TE dimension can be a limiting factor, as the maximum time increase the distance between TEs becomes smaller. To avoid this problem it is possible to increase TE dimensionality or set a reasonable maximum time. 

The main pros of using TEs can be summarized as:

\begin{itemize}
    \item Do not need any optimizable parameter, making it a model-free choice to deal with irregularity. 
    \item Time delta can be linearly computed between two TEs, possibly improving long term dependencies recognition.
    \item All TEs have the same norm, avoiding big values as it is possible to happen with time delta descriptors when interval between observations are big.
\end{itemize}

\section{Experiments}

We evaluate the proposed algorithms on two benchmark tasks: in-hospital mortality and
length of stay  prediction. Booth tasks with the publicly available MIMIC-III dataset \citep{johnson2016mimic}. The following section we will briefly describe the data acquisition and prepossessing used, followed by the test results and discussion.

\subsection{Dataset and training details}

To assess the method performance we used the MIMIC-III benchmark dataset following the benchmark defined by \cite{harutyunyan2017multitask, harutyunyan2019multitask}. With the available code we extracted sequences from in-hospital stays with first $48$ hours and split into training and testing set. This results in a dataset with $17,903$ training samples and $3,236$ test samples for in-hospital mortality after 48 hours task and $35,344/6,225$ for length of stay after 24 hours.

The dataset contains 18 variables with real values and five categorical. We did our own normalization of real variables to zero mean and unit variance, categorical variables are represented with one-hot encoding. At the length of stay task we also change labels from hours to days to avoid large outputs, to report results we change it back to hours.

To make the dataset even more irregular we removed randomly part of observed test data. By doing this we artificially create bigger time gaps to re-evaluate the models with an increased irregularity.

All models was trained with PyTorch \citep{paszke2017automatic} on a P100, with batch size of 100 and AdamW \citep{loshchilov2018decoupled} optimizer with amsgrad \citep{reddi2019convergence}. We performed a five fold cross-validation with 10 runs on each fold. The model with best validation performance (AUC for in-hospital mortality and Mean Absolute Error for length of stay) was selected to compose the average performance for test set. We report the mean and standard error of evaluation measures in test set.

\subsection{Baselines and tests}

To have a baseline we compared TEs primarily with binary masking with time interval indicators, as reported to have a good performance with RNNs in \citep{lipton2016modeling}. It was compared with the proposed method with a regular LSTM \citep{gers1999learning} and a Self-Attentive LSTM \citep{lin2017structured}, as RNNs are reported to achieve best results with the evaluated tasks \citep{lipton2016modeling, shukla2018interpolation, bahadori2019temporal}. 

TEs was tested with dimension ($d_{TE}$) of 32 and maximum time of 48 hours. As binary masking dimension and concatenated TE increase input dimension we adjusted the LSTM hidden size to keep close the number of parameters of as describe at Table \ref{tab:param}. 

All neural models are connected to a 3 layers Multi-Layer Perceptron (MLP) with 32, 32, and 16 neurons. The last layers is a two neurons softmax for in-hospital mortality and one linear, with ReLU, for length of stay.

\begin{table}[]
\centering
\caption{Recurrent models compared and parameter count}
\label{tab:param}
\begin{tabular}{llll}
Model & Alias & h & Nº of params \\ \hline
Vanilla LSTM & LSTM & 34 & 15.5k \\ \hline
LSTM with Binary Masking & BM + LSTM & 22 & 15.5k \\ \hline
LSTM with TE concatenated & catTE + LSTM & 26 & 15.5k \\ \hline
Self-Attentive LSTM & SA-LSTM & 32 & 15.5k \\ \hline
Self-Attentive LSTM with Binary Masking & BM SA-LSTM & 22 & 15.5k \\ \hline
Self-Attentive LSTM with TE added & addTE + SA-LSTM & 32 & 15.5k \\ \hline
\end{tabular}
\end{table}

The Self Attention was implemented as introduced in \citep{lin2017structured} with the only difference of using uni-directional LSTMs. The attention size ($d_{a}$) was 32 and the number of attentions ($r$) was 8. We also used the penalization term of $C=1^{-4}$

As the MIMIC III data are composed by multivariated series and we assume that Time Embeddings (TEs) should not be combined directly. So, we propose to use TEs in two different ways, as additional inputs, replacing missing mask, and as a latent space transformation, by adding TEs to the RNN output hidden state.

\begin{figure}[htb]
\centering
    \includegraphics[width=0.5\linewidth]{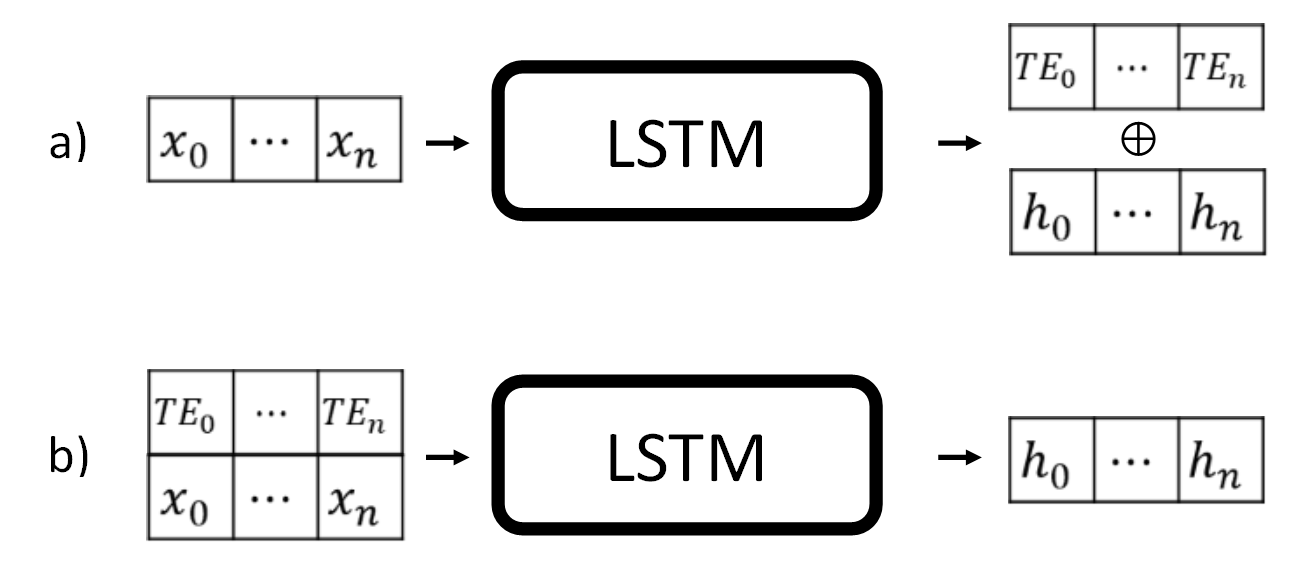}
    \centering
    \captionof{figure}{a) Additive model b) Concatenated model}
\label{sin_discret}
\end{figure}

To have also a baseline of non-recurrent models and assess the TE effect on them, we tested a four layer MLP and Linear/Logistic regression (linear for length of stay and logistic for in-hospital mortality task).

\subsection{Results}

Results for in-hospital mortality shows that self-attention seems to deteriorate the vanilla LSTM performance, but when added the TEs it got improved sufficiently to surpass it and achieve our better average result. 

In the length of stay task TEs achieved better results, especially with bigger gaps at the reduced data test. TEs improved LSTM average error, but a slight worse explained variance, were binary masking had a better performance.

\begin{figure}[]
\centering
\begin{subfigure}{.5\textwidth}
    \includegraphics[width=1.0\linewidth]{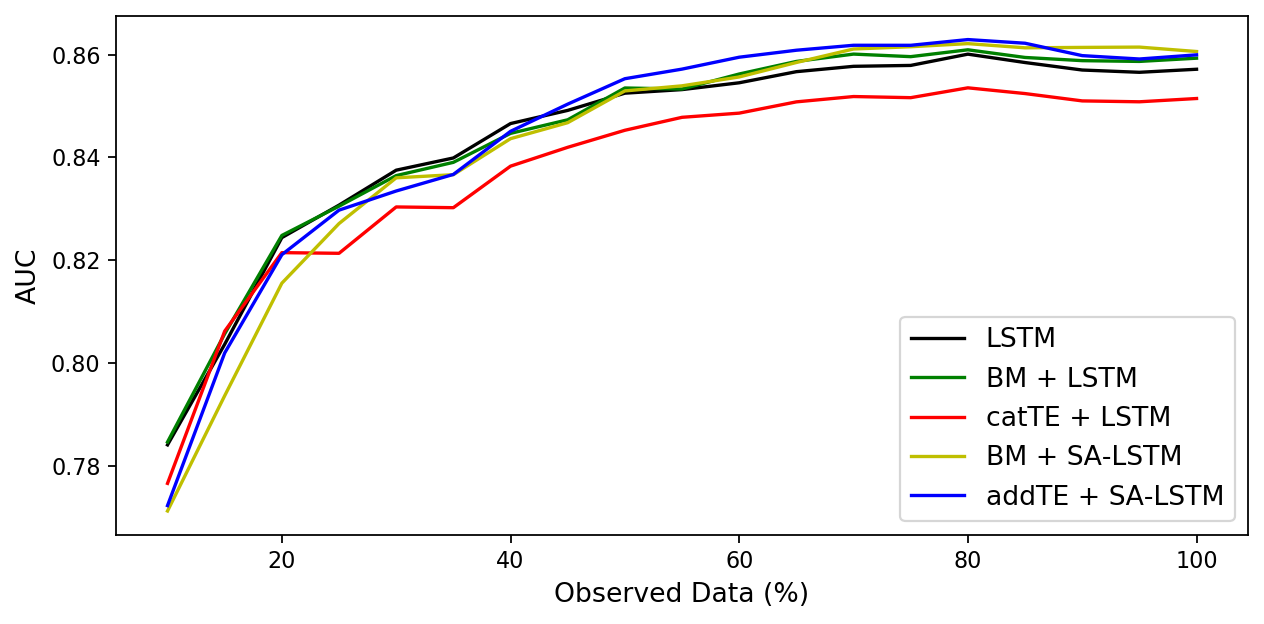}
    \centering
    \caption{AUC-ROC}
    \label{fig:auc}
\end{subfigure}%
\begin{subfigure}{.5\textwidth}
    \includegraphics[width=1.0\linewidth]{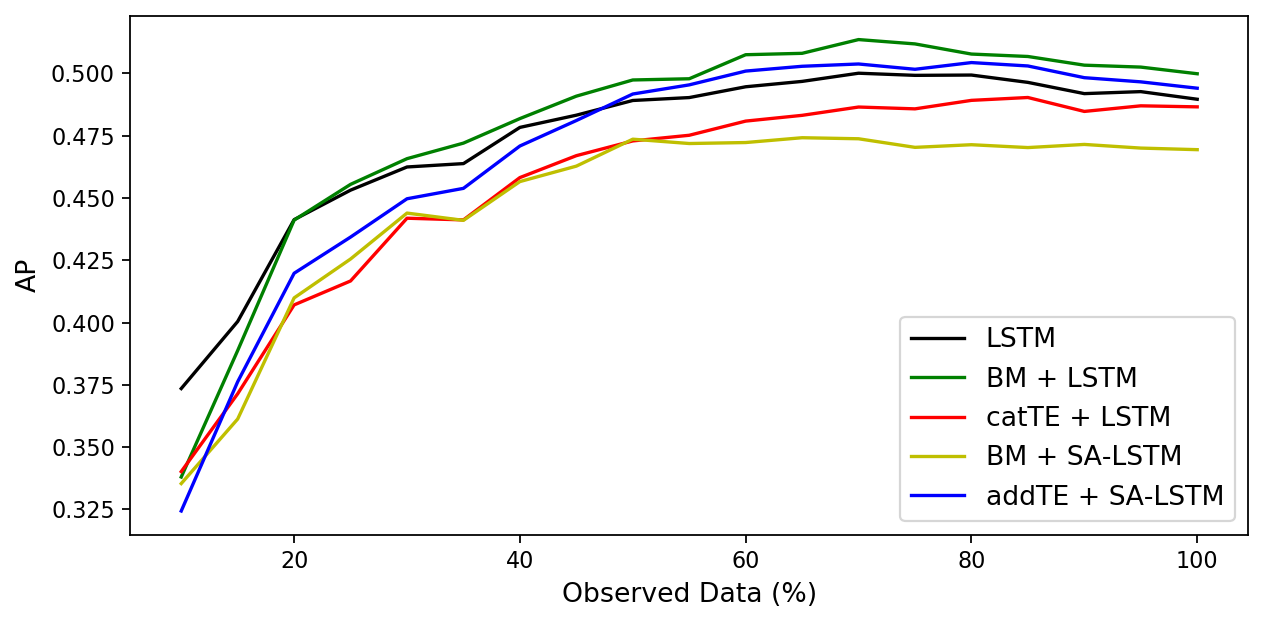}
    \centering
    \caption{Average Precision}
    \label{fig:ap}
\end{subfigure}
\caption{Evaluation of models at in-hospital mortality with observed data from 10\% to 100\%}
\label{fig:observed}
\end{figure}

\begin{figure}[]
\centering
\begin{subfigure}{.5\textwidth}
    \includegraphics[width=1.0\linewidth]{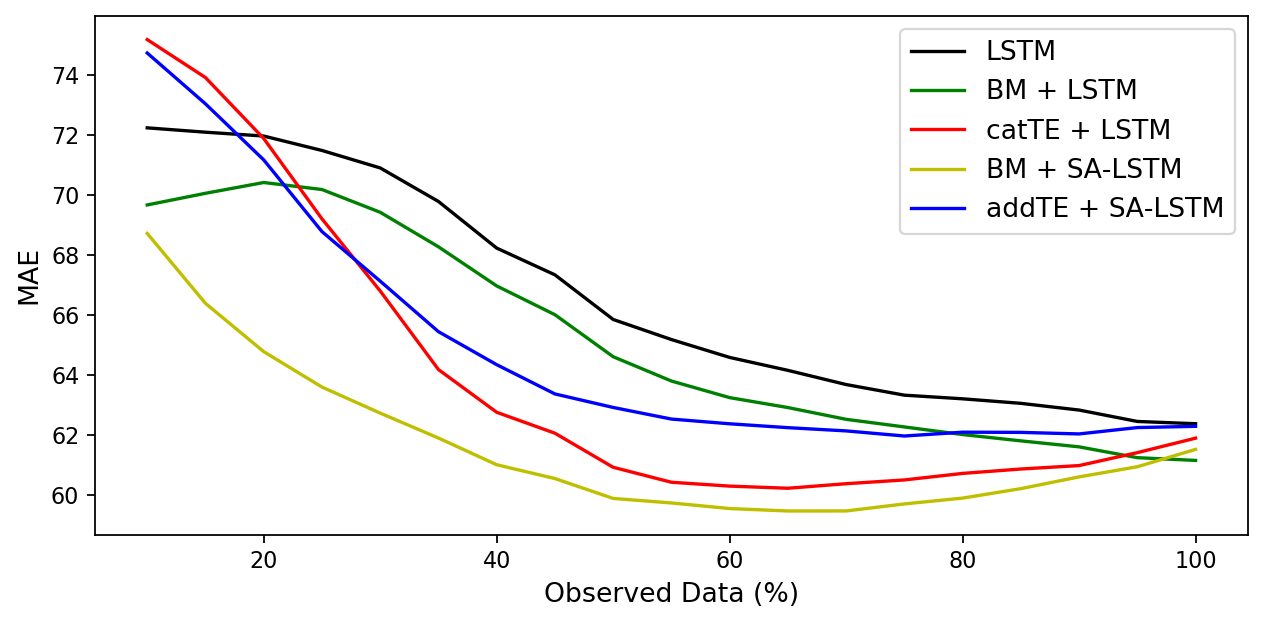}
    \centering
    \caption{Mean Absolute Error}
    \label{fig:mae}
\end{subfigure}%
\begin{subfigure}{.5\textwidth}
    \includegraphics[width=1.0\linewidth]{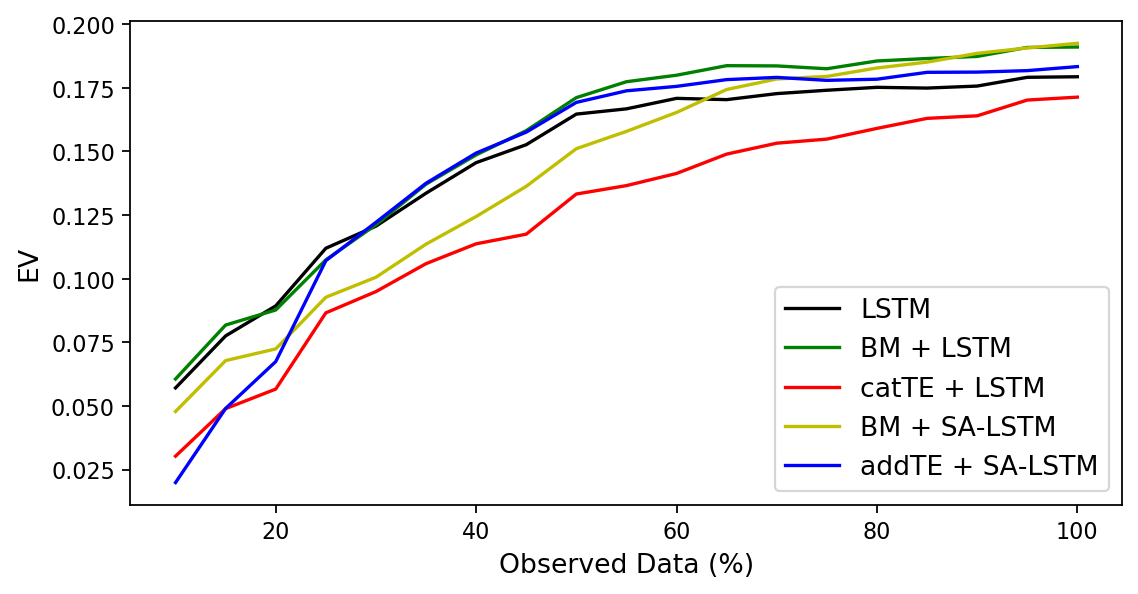}
    \centering
    \caption{Explained Variance}
    \label{fig:ev}
\end{subfigure}
\caption{Evaluation of models at length of stay with observed data from 10\% to 100\%}
\label{fig:observed_los}
\end{figure}

At Figure \ref{fig:observed} and \ref{fig:observed_los} we can see the performance of models when we randomly remove observed data from 100\% to 10\%. With length of stay task the LSTM with TE concatenated have a overall smaller absolute error than vanilla LSTM, being surpassed only by the binary mask. At in-hospital mortality we see a similar performance with TE SA-LSTM and LSTM with binary masking.

With non-recurrent models it is possible to observe how TEs does not rely on recurrence. It improved both linear/logistic regression and MLP.

\begin{table}[]
\centering
\caption{Results of Mean Absolute Error (MAE), Root Mean Squared Error (RMSE) and Explained Variance (EV) for length of stay task and Area Under ROC-Curve (AUC-ROC) and Average Precision (AP) for in-hospital mortality task. 
All results are in \textit{mean(standard deviation)} format}
\label{tab:results}
\resizebox{\textwidth}{!}{%
\begin{tabular}{rllllll}
Model & \multicolumn{3}{c}{24h Length-of-Stay} &  & \multicolumn{2}{c}{48h In-hospital Mortality} \\ \hline
 & MAE & RMSE & EV &  & AUC-ROC & AP \\ \hline
Lin.R/Log.R & 65.829(2.001) & 133.178(2.777) & 0.040(0.034) &  & 0.783(0.003) & 0.357(0.357) \\
BM + Lin.R/Log.R & 68.343(2.453) & 127.579(2.044) & 0.104(0.029) &  & 0.804(0.003) & 0.369(0.369)) \\
catTE + Lin.R/Log.R & 69.585(0.870) & 129.484(1.220) & 0.077(0.017) &  & 0.794(0.005) & 0.359(0.359) \\ \hline
MLP & 64.983(1.695) & 137.949(14.440) & -0.053(0.236) &  & 0.807(0.002) & 0.374(0.374)) \\
BM + MLP & 63.181(3.120) & 125.888(0.781) & 0.133(0.007) &  & 0.804(0.004) & 0.365(0.365)) \\
catTE + MLP & 64.154(1.734) & 129.259(2.115) & 0.090(0.019) &  & 0.805(0.003) & 0.355(0.355)) \\ \hline
LSTM & 62.597(0.957) & 122.284(0.266) & 0.174(0.004) &  & 0.855(0.002) & 0.481(0.481) \\
BM + LSTM & \textbf{61.666(0.702)} & 121.899(0.442) & 0.180(0.006) &  & 0.854(0.004) & \textbf{0.485(0.485)} \\
catTE + LSTM & 62.354(0.686) & 123.253(0.155) & 0.164(0.002)) &  & 0.846(0.004) & 0.476(0.476) \\ \hline
SA-LSTM & 63.353(0.715) & 122.976(0.215) & 0.167(0.003) &  & 0.851(0.005) & 0.448(0.448) \\
BM + SA-LSTM & 61.955(0.460) & \textbf{121.667(0.089)} & \textbf{0.185(0.001)} &  & 0.854(0.006) & 0.454(0.454)) \\
addTE + SA-LSTM & 62.591(0.321) & 122.325(0.475) & 0.173(0.006) &  & \textbf{0.856(0.003)} & 0.482(0.482) \\ \hline
\end{tabular}%
}
\end{table}

\section{Conclusions}

This paper propose a novel method to represent hour time or dates as dense vectors to improve irregularly sampled time series. It was evaluated with two different approaches and evaluated in two tasks from the MIMIC III dataset. Our method showed some improvement with most models tested, including recurrent neural networks and classic machine learning methods.

Despite being outperformed by binary masking in some tests we believe TEs can still be an viable option. Specially to very irregular time series and high dimensional data, were TEs can be applied by addition without increasing the input dimensionality.

\section{Future Work}

We see a promising future for the method proposed. We expect to extend it to improve other types of irregular time-continuous data and also evaluate how can TE improve recent models proposed for irregularly time series, like the GRU-D \citep{che2018recurrent}, interpolation networks \citep{shukla2018interpolation} and Temporal-Clustering Regularization \citep{bahadori2019temporal}. The code for TEs results reported will be publicly available in the future.

\bibliography{iclr2020_conference}
\bibliographystyle{iclr2020_conference}

%\appendix
%\section{Appendix}
%You may include other additional sections here. 

\end{document}